\documentclass[journal]{IEEEtran}
\usepackage{amsmath}
\usepackage{amssymb}
\usepackage{algorithm}
\usepackage{algorithmic}
\usepackage{comment}
\usepackage{mathtools}
\usepackage{graphicx}
\usepackage{tabularx}
\usepackage{adjustbox}
\usepackage{subcaption}
\usepackage{varwidth}
\usepackage{xcolor}
\usepackage{booktabs}
\usepackage{pgfplotstable}
\usepackage{lineno}
\usepackage{hyperref}
\modulolinenumbers[5]
\definecolor{RevCol}{RGB}{255,0,0}

\begin{document}

\title{Km-scale dynamical downscaling through conformalized latent diffusion models}

\author{Alessandro Brusaferri and Andrea Ballarino%
\thanks{Alessandro Brusaferri (corresponding author) and Andrea Ballarino are with CNR, Institute of Intelligent Industrial Technologies and Systems for Advanced Manufacturing, via A. Corti 12, Milan, Italy (e-mail: alessandro.brusaferri@cnr.it).}%
}

\maketitle

\begin{abstract}
Dynamical downscaling is crucial for deriving high-resolution meteorological fields from coarse-scale simulations, enabling detailed analysis for critical applications such as weather forecasting and renewable energy modeling. Generative Diffusion models (DMs) have recently emerged as powerful data-driven tools for this task, offering reconstruction fidelity and more scalable sampling supporting uncertainty quantification. However, DMs lack finite-sample guarantees against overconfident predictions, resulting in miscalibrated grid-point-level uncertainty estimates hindering their reliability in operational contexts.
In this work, we tackle this issue by augmenting the downscaling pipeline with a conformal prediction framework. Specifically, the DM’s samples are post-processed to derive conditional quantile estimates, incorporated into a conformalized quantile regression procedure targeting locally adaptive prediction intervals with finite-sample marginal validity. The proposed approach is evaluated on ERA5 reanalysis data over Italy, downscaled to a 2-km grid. Results demonstrate grid-point-level uncertainty estimates with markedly improved coverage and stable probabilistic scores relative to the DM baseline, highlighting the potential of conformalized generative models for more trustworthy probabilistic downscaling to high-resolution meteorological fields.
\end{abstract}


\section{Introduction}
\label{Intro}
Accurate and computationally efficient dynamical downscaling of meteorological fields to kilometer-scale resolutions remains a central challenge in weather and climate science \cite{xu2019dynamical}. Fine-grained predictions capturing localized atmospheric phenomena and extreme events provide critical information for a broad range of downstream applications, including renewable energy forecasting, disaster risk management, and agricultural planning \cite{Merizzi2024},\cite{Craig.etal2022}.
Operational dynamical downscaling relies on regional climate models that numerically solve the physical equations governing the atmosphere, starting from coarser-resolution conditions generated by global circulation models. While this approach offers physically consistent and high-fidelity representations of local climate features, the substantial computational expense involved significantly limits the size of output ensembles, thus constraining comprehensive uncertainty quantification \cite{Rampal24}.

To overcome this barrier, statistical downscaling methods have been widely adopted as efficient surrogates by inferring empirical relationships between large-scale climate variables and local-scale behavior. Traditional approaches in this class include quantile mapping \cite{PanofskyBrier1968}, multilinear regression methods \cite{Sharifi19} and analog ensemble \cite{Sperati24}.
Subsequently, deep learning techniques have garnered substantial attention from the community owing to their enhanced capacity to model complex spatial dependencies, nonlinear dynamics, and the intricate relationships embedded within the conditioning set \cite{gmd-13-2109-2020}. In this context, research has progressed from deterministic mapping approaches, such as convolutional and attention-based neural networks, toward generative architectures capable of probabilistic downscaling that address the intrinsic stochasticity of fine-scale atmospheric processes. Early generative adversarial network (GAN)-based techniques (see e.g., \cite{Wang-2021}) produced spatially coherent high-resolution outputs. However, GANs suffer from training instability, mode collapse, limited sample diversity, and demanding tricky tuning and architectural choices. More recently, diffusion models (DMs) have emerged as a promising alternative for complex conditional generation, offering more stable training through a likelihood-based objective and better approximation of the target conditional distribution (see, e.g., \cite{Rampal24}, for a detailed review).

Building upon this framework, \cite{Mardani2025} introduced residual corrective approaches, demonstrating that combining a deterministic model—designed to capture the dominant large-scale high-resolution dynamics—with a diffusion model specifically trained to represent localized stochastic variability leads to improved reconstructions. Nevertheless, full image-space diffusion computations and memory consumption grow significantly with increasing resolution, creating a critical computational bottleneck. To address this issue, \cite{gmd-18-2051-2025} deployed Latent Diffusion Models \cite{LDM_22} within the residual corrective architecture, exploiting a compressed latent space to enable more scalable training and sampling of high-resolution outputs. The resulting model outperforms both traditional and GAN-based downscaling methods, demonstrating superior performance in reconstructing fine-scale details and maintaining accuracy in frequency and spatial error distributions.

Despite these compelling enhancements, the reliability of uncertainty estimates in DM-based downscaling techniques remains critical. In particular, generated scenarios have been shown to be potentially under-dispersive - i.e, their ensemble spread is overly narrow relative to the actual error - thus undermining probabilistic predictions trustworthiness. In fact, obtaining robust uncertainty calibration in diffusion models constitutes a pivotal research problem in the broader machine learning context \cite{10.5555/3618408.3619822}. For the target dynamical downscaling purpose, the aim is to achieve reliable grid-point-level prediction uncertainty estimates matching local observed error levels, properly representing the range of possible values location-wise with varying levels of confidence.

In this work, we aim to tackle this issue by exploring Conformal Prediction (CP). Specifically, we extend the data-driven downscaling framework by introducing an efficient asymmetric conformalized quantile regression approach, deployed in a grid-point-wise manner to support local prediction interval reliability (i.e., sample-wise efficiency) beyond the marginal coverage guarantees obtained by applying CP on deterministic predictors. To this end, residual corrective LDM samples serve as first-stage proxies to derive heuristic notions of uncertainty as conditional quantile estimates, which are then conformalized to produce calibrated intervals tailored to each grid point. The experiments are conducted on open datasets, including dynamical downscaling samples generated by the COSMO-CLM regional climate model driven by ERA5 reanalysis. The focus is on high-resolution fields of 2-meter temperature and 10-meter horizontal wind components over Italy, increasing the native resolution to approximately 2.2 km.

\section{Methods}
\label{Methods}
\subsection{Probabilistic AI-drive downscaling problem}
We focus on AI-driven dynamical downscaling, aiming to learn a data-driven mapping from low-resolution global climate model outputs to high-resolution regional fields. Let $Y \in \mathbb{R}^{H \times W \times v}$ denote the coarse-scale climate predictors, where $H$, $W$ represent spatial dimensions and $v$ the number of climate variables. We note that the conditioning set can include additional exogenous features at both low and high resolutions, following appropriate projections and concatenation. These additional features are made implicit hereafter to simplify notation. Similarly, let $X \in \mathbb{R}^{h \times w \times v}$ be the corresponding fine-scale, high-resolution climate targets with spatial dimensions $h \gg H$ and $w \gg W$. The objective is to train a parameterized model $f_{\theta}$ using sample data pairs $(Y, X) \sim \mathcal{D}$ such that $f_{\theta}: Y \mapsto \hat{X}$, with $\hat{X} \approx X$, effectively serving as a computationally efficient surrogate approximating the nonlinear dynamical downscaling process performed by traditional regional climate models.

The dynamical downscaling emulation task involves an inherently ill-posed mapping, common in image-to-image inverse problems. 
Hence, we target a probabilistic formulation aimed to identify the overall conditional distribution $\mathbb{P}(X|Y)$ describing the range of plausible downscaled regional climate realizations given the low-resolution inputs, better capturing the intrinsic stochasticity of the downscaling emulation.
\subsection{Conformal prediction}
Conformal Prediction (CP) provides a flexible framework for constructing prediction sets that guarantee marginal coverage for any out-of-sample test point $(Y_k, X_k) \sim \mathcal{D}_{test}$:
\begin{equation}
\mathbb{P}\big(X_k \in \mathcal{C}_{1-\alpha}(Y_k)\big) \geq 1 - \alpha,
\end{equation}
where $\alpha$ denotes the target error tolerance level \cite{angelopoulos2022gentle}. In this formulation, we follow the convention of downscaling for consistency.
Achieving full coverage is trivial if efficiency is not considered; for example, in continuous settings choosing \(\mathcal{C}_{1-\alpha}(Y_k) = \mathbb{R}\) yields \(\mathbb{P}(\cdot) = 1\). Therefore, the primary goal is to produce sharp prediction sets that closely match the desired coverage level, i.e., $\mathbb{P}\big(X_k \in \mathcal{C}_{1-\alpha}(Y_k)\big) \approx 1 - \alpha$

Split conformal prediction constructs prediction intervals by extracting a dedicated calibration subset \((Y_c, X_c) \sim \mathcal{D}_c\), distinct from the data used to fit the model \(f_\theta\), where conformity scores quantify prediction uncertainty by measuring agreement with observed outcomes (see e.g., \cite{Fontana2023Conformal} for a recent review). A typical conformity score for regression problems is the absolute residual $\mathcal{S}_c = |X_c - f_\theta(Y_c)|$.
By ranking these conformity scores and calculating empirical quantiles over the \(n\) calibration samples, under exchangeability assumptions, it follows that for any test sample \(k\):
\begin{align}
    &\mathbb{P}(|X_k - f_\theta(Y_k)| \leq \mathcal{S}_{(\lceil(n+1)(1-\alpha) \rceil)}) \nonumber \\
    &=\mathbb{P}(X_k \in \underbrace{f_\theta(Y_k) \pm \mathcal{S}_{(\lceil(n+1)(1-\alpha) \rceil)}}_{\mathcal{C}_{1-\alpha}(Y_k)})\\
    &=\frac{\lceil(n+1)(1-\alpha) \rceil}{n} \geq 1-\alpha
    \label{eq1}
\end{align}
This finite-sample guarantee holds marginally over the data-generating distribution.
An upper bound is derived by the random tie-breaking rule to handle equality in rankings \cite{b2}, i.e.,
$\mathbb{P}\big(X_k \in \mathcal{C}_{1-\alpha}(Y_k)\big) \leq 1 - \alpha + \frac{1}{n+1}$.

As indicated by the above expressions, conventional split CP based on absolute conformity scores from deterministic predictions produces symmetric prediction intervals with limited variations across the input space, leading to conservative uncertainty quantification \cite{angelopoulos2024theoreticalfoundationsconformalprediction}. Such average-width bands provide limited insight into sample-specific uncertainty for practical dynamical downscaling applications, for example, failing to appropriately characterize the grid-wise uncertainty magnitude depending on the current conditioning variables.
Still, since exact conditional coverage $\mathbb{P}(X_k \in \mathcal{C}_{1-\alpha}(Y_k)|Y_k = \nu) \geq 1 - \alpha$ for almost all $\nu$, is unattainable under finite-sample and distribution-free assumptions, the practical objective is to approximate local adaptivity in prediction intervals while yielding target marginal coverage.

\subsection{Improved local adaptivity by conformalized quantiles}
To achieve improved local adaptivity with grid-wise coverage guarantees, we develop a conformalized quantile regression-based approach within the dynamical downscaling framework. 
Specifically, we leverage the asymmetric conformity score computations introduced by \cite{NEURIPS2019_5103c358} for regression tasks, here applied grid-wise as follows:
\begin{align}
  \mathcal{C}^{(i,j)}_{1-\alpha}(Y_{k})=[&{q}^{(i,j)}_{\alpha/2}(Y_k)-{l}^{(i,j)}_{1-\alpha/2}(\mathcal{D}_{c}), \\ 
  &{q}^{(i,j)}_{1-\alpha/2}(Y_k)+{u}^{(i,j)}_{1-\alpha/2}(\mathcal{D}_{c})]   \nonumber
\end{align}
for $i=1,...,h, j=1,...,w$, with $\mathcal{C}^{(i,j)}_{1-\alpha}(Y_{k})$ defining the (1-$\alpha$)-level prediction interval for the grid point $(i,j)$ in the high-resolution target image given the conditioning input $Y_k$. 
$\small{{l}^{(i,j)}_{1-\alpha/2}\small}$ and ${u}^{(i,j)}_{1-\alpha/2}$ depict the (1-$\alpha/2$)-th empirical quantiles of ${q}^{(i,j)}_{\alpha/2}(Y_c)-X_c^{(i,j)}$ and $X_c^{(i,j)}-{q}^{(i,j)}_{1-\alpha/2}(Y_c)$ respectively, computed on the calibration subset $c \in \mathcal{D}_{c}$. 
$\{{q}^{(i,j)}_{{\gamma}}(Y_k)\}_{{\gamma} \in \Gamma}, \: {q}^{(i,j)}_{{\gamma}}(Y_k) \leq {q}^{(i,j)}_{{\gamma}'}(Y_k) \; \forall {\gamma} <{\gamma}'$ represent the discrete set $\Gamma$ of the conditional prediction quantiles derived from a set of residual corrective diffusion model samples (see section \ref{DM_par}).
The bounds of the prediction intervals corresponding to specified coverage levels are derived from the predicted quantiles within the set $\Gamma$, which is defined to cover balanced quantile pairs such as deciles, percentiles, etc. The sample size must be sufficiently large to accurately estimate the quantiles at the desired coverage levels.

Overall, applying CP to models that provide a conditioned heuristic measure of uncertainty - such as diffusion models asymptotically approximating the latent distribution - yields more reliable, sample-wise uncertainty quantification. This approach combines the reconstruction capabilities of residual corrective diffusion models for dynamical downscaling with the calibration support of CP. Furthermore, it extends grid-level normalized conformal inference methods (see e.g., \cite{gopakumar2024valid}), which are limited by producing symmetric bounds around point predictions regardless of the underlying distribution shape and suffer from potential interval inflation and systematic underestimation issues \cite{angelopoulos2022gentle}.
We note that pixel-wise conformity score computation does not inherently account for coverage dependencies among adjacent points, whose approximate characterization is left to the backbone learning process \cite{gopakumar2024valid}. In fact, this can be reformulated as a Conformal Risk Control problem with grid-wise coverage targets and thresholds $\lambda^{(i,j)}$ (see e.g., \cite{angelopoulos2025conformalriskcontrol}, \cite{10.5555/3618408.3619822}). Exploring further multivariate CP extensions (such as in \cite{dheur2025unifiedcomparativestudygeneralized}) within the dynamical downscaling framework represents a promising direction for future works.

\subsection{Residual corrective grid-level quantiles estimation}
\label{DM_par}
As introduced in Section \ref{Intro}, we deploy a residual corrective architectural form combining a deterministic mapping, intended to capture the dominant large-scale dynamics, with a latent diffusion model-based component devoted to localized stochastic variability. The exploration of further architectural designs within the CP framework is left to future studies.

The former is first trained, by minimizing a mean squared reconstruction error, producing a baseline high-resolution estimate $\hat{X}_{\text{d}} = U_{\phi}(Y)$ leading to the residual $R = X - \hat{X}_{\text{d}}$.
$R$ is then projected sample-wise into a lower-dimensional latent space $Z_0$ using a parametric Variational Autoencoder (VAE) consisting of an encoder and decoder 
$Z_0 = E_{\psi}(R), \hat{R} = D_{\phi}(Z_0)$,
learned through a combined mean absolute reconstruction error and Kullback-Leibler divergence of the latent states with respect to a multivariate standard Normal prior.

Diffusion models are built to support sampling from the latent conditional distribution by learning to revert a forward diffusion process gradually destroying the input data structure by adding noise.
Formally, the forward diffusion process is modeled by a Markov chain of transitions:
\begin{equation}
q(Z_t \mid Z_{t-1}) = \mathcal{N}\left(Z_t; \sqrt{1 - \beta_t} Z_{t-1}, \beta_t I \right).
\end{equation}
with a variance schedule \(\{\beta_t\}_{t=1}^T\).
Then, a neural network including a conditioning component is trained to approximate the score function for the reverse diffusion:
\begin{equation}
s_{\theta}(Z_t, t \mid Y) \approx \nabla_{Z_t} \log p(Z_t \mid Y)   
\end{equation}
enabling generation of conditioned latent residual samples:
\begin{equation}
p_{\theta}(Z_{t-1}|Z_t, Y) := \mathcal{N}\left(
Z_{t-1}; Z_t + \sigma_t^2 \nabla_{Z_t} \log p(Z_t|), \sigma_t^2 I
\right) \nonumber
\end{equation}
At inference, starting from random noise instances \(Z_T \sim \mathcal{N}(0, I)\), the reverse diffusion process generates \(M\) independent latent residual samples:
\begin{equation}
\{\hat{Z}_0^{(m)}\}_{m=1}^M \sim p_{\theta}(Z_0 \mid Y).
\end{equation}
These samples are decoded back to residual space:
\begin{equation}
\hat{R}^{(m)} = D_{\phi}(\hat{Z}_0^{(m)}), \quad m=1,\dots,M
\end{equation}
and combined with the deterministic prediction to produce final downscaled samples $\hat{X}^{(m)} = \hat{X}_{d} + \hat{R}^{(m)}$.
Pixel-wise predictive quantiles at level $\gamma \in \Gamma$ are estimated empirically from the high-resolution regional level fields ensemble:
\begin{equation}
q_{\gamma}^{(i,j)}(Y_k) = \inf \left\{ x : \frac{1}{M} \sum_{m=1}^M \mathbf{1}\left( \hat{X}^{(m)}_{(i,j)} \leq x \right) \geq \gamma \right\}
\end{equation}
where \(\hat{X}^{(m)}_{(i,j)}\) is the value of sample \(m\) at grid-point \((i,j)\) in the target image.
These are computed over a couple set of discrete quantiles, leading to PIs of increasing coverage degrees reflecting the range of plausible downscaled values at each spatial location injected in the CP procedure. Specific configurations tailored to each application are defined, as described in the following section. 

\section{Experiments and Results}
\label{Results}
\begin{figure*}[t!]
  \centering
  \includegraphics[width=\textwidth]{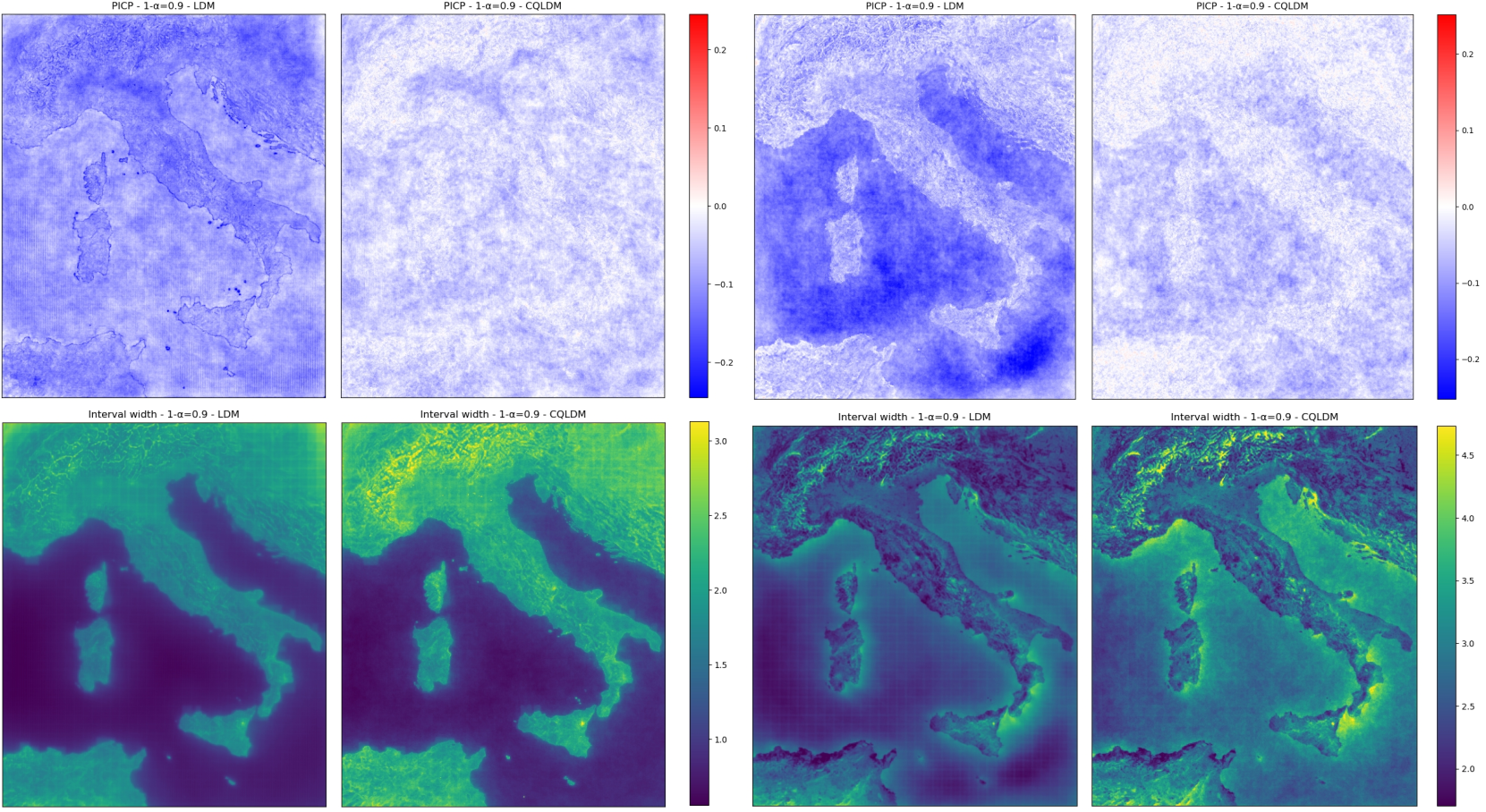}
  \caption{Grid-wise PICP and PI width for coverage level 1-$\alpha$=0.9. (l)- 2mT, (r)-WS}
  \label{fig:grid_picp_W}
\end{figure*}
To evaluate the proposed approach, we utilize the open repository provided by \cite{gmd-18-2051-2025}, which offers a robust framework for comparative evaluations of challenging dynamical downscaling tasks. We summarize the major components hereafter, referring interested readers to \cite{gmd-18-2051-2025} for details.

\subsection{Case studies}
The benchmark problem targets high-resolution hourly meteorological data over Italy, obtained by dynamically downscaling ERA5 reanalyses (25 km resolution) to 2.2 km using the COSMO5.0\_CLM9 model (VHR-REA\_IT CCLM). This setup requires the downscaling model to address a more complex challenge than simple super-resolution, as it must correct large-scale biases and generate fine-scale phenomena absent at coarse resolution. ERA5 fields are interpolated onto a 16 km grid, while the high-resolution outputs are on a 2 km grid, resulting in an 8$\times$ downscaling factor over a domain covering 35--48$^\circ$ N latitude and 5--20$^\circ$ E longitude (i.e., from 72×86 pixels to 576×672 pixels). This region includes diverse topography, such as mountains and plains, increasing the complexity of the downscaling task. 
The focus is on producing high-resolution 2 m temperature and 10 m wind component fields, which present specific downscaling difficulties due to their distinct physical characteristics. 14 input channels are used, comprising temperature, wind components at two levels, pressure, sea surface temperature, snow depth, dew-point temperature, solar radiation, humidity, and precipitation. High-resolution static data like elevation, land cover, and latitude are also incorporated.

The dataset spans hourly data from 2000 to 2020 and is randomly split into training, validation, and testing subsets with uniform temporal coverage. The out-of-sample test set comprises 5\% of the data, designed to optimize evaluation efficiency for resource-intensive diffusion models. For the present study, we extracted a random subset of 1,460 samples from the original test set and split it evenly into two subsets of 730 samples each. The first subset serves as the calibration set for calculating conformity scores, while the second subset constitutes the final test set used to assess performance. The calibration set includes observations occurring before the first testing date. These subsets were sized to balance the growing computational cost while involving heterogeneous downscaling conditions, as generating multiple samples from the diffusion model for both calibration and testing—necessary for obtaining the related prediction quantiles—requires substantially more processing time compared to the original study, which performed single-sample evaluations using reconstruction metrics. 
\subsection{Experimental setup}
The deployed Latend Diffusion Model (LDM) architecture follows the design specifications in \cite{gmd-18-2051-2025}. The deterministic U-Net consists of an encoder with 4 blocks, each containing two 2D convolutional layers followed by batch normalization, ReLU activations, and strided max-pooling. The decoder replicates the encoder’s structure, employing transposed 2D convolutional layers and upsampling, along with skip connections. The VAE performing residual projections stacks a 2D convolutional encoder and decoder composed of blocks that integrate ResNet residual units with convolutional layers for downsampling and upsampling. Three hierarchical levels are employed, each halving the spatial resolution, while constraining the channel dimension to a maximum of 32 times the number of input target variables.
The denoiser component, which performs diffusion in the latent space, consists of 2 paired downsampling and upsampling blocks including cross-AFNO (Adaptive Fourier Neural Operator) \cite{leinonen2023latentdiffusionmodelsgenerative} and 2D residual layers. A context encoder preprocesses low-resolution predictors and high-resolution static data, embedding them for each level of the denoiser architecture. After proper encoding and projection via 2D convolutions, these are passed through a sequence of four 2D AFNO layers, followed by targeted combinations of average pooling and 2D residual blocks.

A dedicated residual network chain has been developed for each target variables, i.e, the 2m temperature (2mT) and 10m horizontal wind components (WS).
The deterministic U-Nets were initially trained using Adam with a learning rate of 1e-3 and a batch size of 16. Subsequently, the VAEs were trained on random 512\(\times\)512 pixel patches of high-resolution target variables, employing AdamW with a base learning rate of 1e-3 and a KL weight factor of 0.01. Finally, the conditioner and denoiser were trained jointly by minimizing the mean squared error in a v-prediction parameterization, using random patches of 64\(\times\)64 pixels for low-resolution features and 512\(\times\)512 pixels for static features. The AdamW learning rate was set to 1e-4, with a 0.25-factor scheduler to reduce rate on plateau and exponential moving average utilities. To manage memory constraints, batch sizes were set to 8 and 4 for the 2mT and WS components, respectively. The diffusion noise schedule follows a linear pattern over 1000 discrete steps in the range [1e-4, 2e-2]. Across all LDM architectural components, maximum epochs is set to 100, with early stopping patience of 10.

To assess the probabilistic performance on the test sets, we included metrics capturing both calibration and sharpness, essential for reliable uncertainty quantification. The calibration of the obtained intervals was primarily evaluated using the standard Prediction Interval Coverage Probability (PICP). Proper scoring rules, namely the Interval Score (IS, aka Winkler) and the Quantile Score (QS, aka Pinball), were then employed. Formally, grid-wise IS and QS are defined as:
\begin{align}
\small
\text{IS}_{1-\alpha}^{(i,j)} &=\begin{cases}
        \delta_{1-\alpha}^{(i,j)}, \quad \quad \mathrm{X}^{(i,j)} \in [\mathcal{L}_{1-\alpha}^{(i,j)}, \mathcal{U}_{1-\alpha}^{(i,j)}]\\
        \delta_{1-\alpha}^{(i,j)} + \frac{2}{\alpha}(\mathcal{L}_{1-\alpha}^{(i,j)}-\mathrm{X}^{(i,j)}), \quad \mathrm{X}^{(i,j)} < \mathcal{L}_{1-\alpha}^{(i,j)}\\
        \delta_{1-\alpha}^{(i,j)} + \frac{2}{\alpha}(\mathrm{X}^{(i,j)}-\mathcal{U}_{1-\alpha}^{(i,j)}), \quad \mathrm{X}^{(i,j)} > \mathcal{U}_{1-\alpha}^{(i,j)}\\
    \end{cases}
    \nonumber\\
\text{QS}_{\gamma}^{(i,j)}&=(\mathrm{X}^{(i,j)}-\hat{q}_{\gamma}^{(i,j)})\gamma \mathrm{1}\{\mathrm{X}^{(i,j)}>\hat{q}_{\gamma}^{(i,j)}\}\nonumber \\ 
&+ (\hat{q}_{\gamma}^{(i,j)}-\mathrm{X}^{(i,j)})(1-\gamma) \mathrm{1}\{\mathrm{X}^{(i,j)}\leq \hat{q}_{\gamma}^{(i,j)}\} \nonumber
\end{align}
with $\hat{q}_{\gamma}^{(i,j)}$ the predicted quantile and $\delta_{1-\alpha}^{(i,j)}=\mathcal{U}_{1-\alpha}^{(i,j)}-\mathcal{L}_{1-\alpha}^{(i,j)}$ the (1-$\alpha$)-PI width for grid point (i,j). 
The first case in IS rewards narrow PIs (i.e., sharpness), while the others penalizes the occurrence of test observations outside the PI.

The IS balances interval sharpness and coverage by rewarding narrow intervals that contain the true outcome, while the QS evaluates the accuracy of quantile forecasts by penalizing asymmetric errors. Additionally, the average prediction interval width (IW) was analyzed to assess sharpness.
The target PI coverage levels have been defined in the discrete set \(1-\alpha \in \{0.2, 0.4, 0.6, 0.8\}\), yielding an output quantile set 
$\gamma \in \Gamma = \{ \gamma|\gamma = 0.05+k, k=0,0.1,...,0.9 \}$ aimed to characterize both central tendencies and tail behavior of the predictive distribution, enabling assessment of PIs across varying risk levels. For the scope of the present work, the probabilistic forecasting metrics are computed at grid-level to assess local calibration and sharpness, with subsequent averaging to provide a summary of overall map-level performance.
\subsection{Results}
\begin{figure}[t!]
\begin{center}
\includegraphics[width=0.96\linewidth]{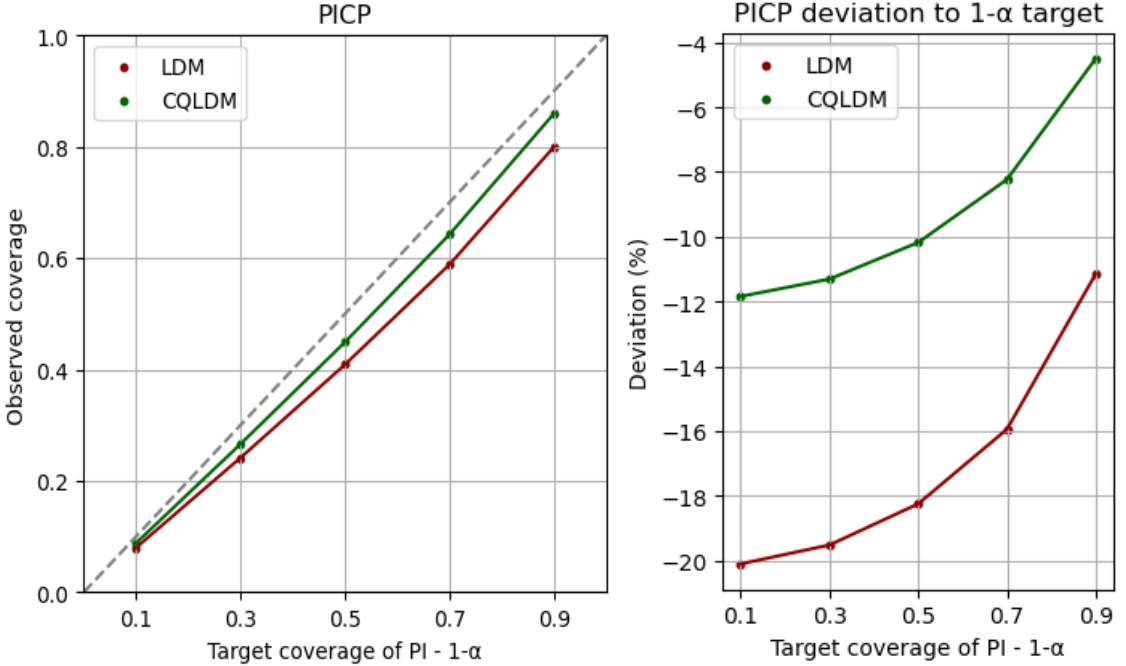}
\end{center}
\caption{2mT average PICP and \% Deviation}
\label{PICP_2mT}
\end{figure}
\begin{figure}[t!]
\begin{center}
\includegraphics[width=0.96\linewidth]{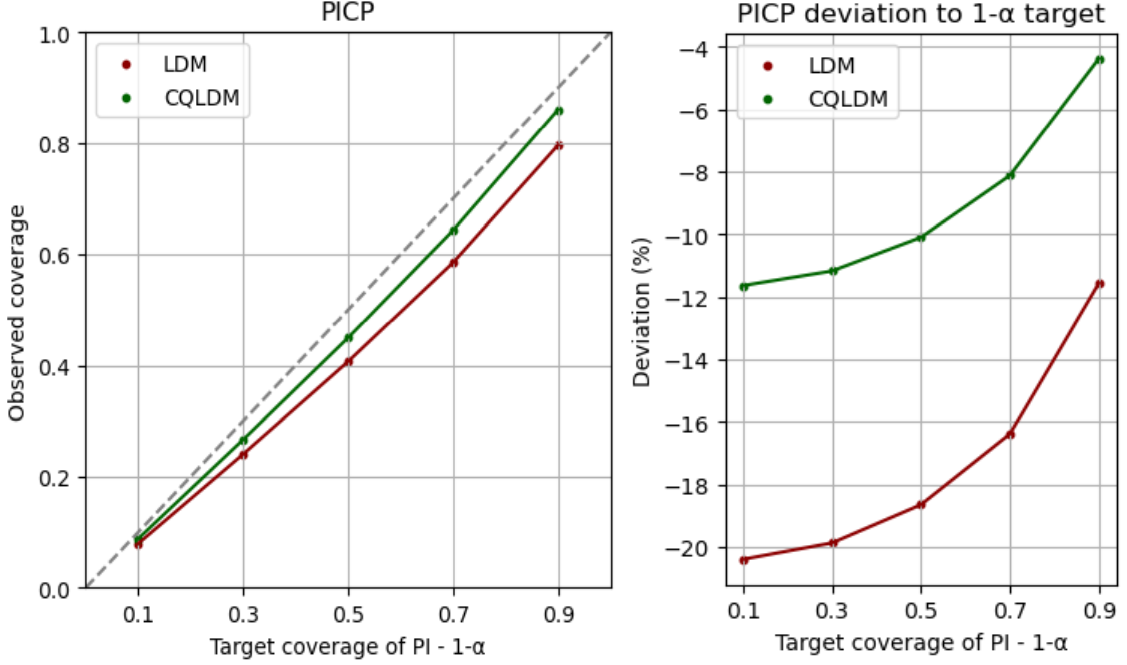}
\end{center}
\caption{WS average PICP and \% Deviation}
\label{PICP_WS}
\end{figure}

\begin{table}[t!]
\caption{Interval Scores and related Interval Widths}
\label{Tab_IS_IW}
\begin{center}
\small
\begin{tabular}{llllll}
\hline
\multicolumn{6}{c}{\bf{2mT}} \\
\hline
\bf{IS$_{1-\alpha}$} &0.1 &0.3 &0.5 &0.7 &0.9 
\\ \hline
LDM &0.923  &1.116 &1.357 &1.706 &2.479 \\ 
CQLDM &0.921 &1.112 &1.345 &1.672 &2.322 \\ \hline
\bf{IW$_{1-\alpha}$} &0.1 &0.3 &0.5 &0.7 &0.9 
\\ \hline
LDM &0.101 &0.308 &0.539 &0.830 &1.319 \\ 
CQLDM &0.111 &0.342 &0.603 &0.937 &1.537 \\ \hline
\hline
\multicolumn{6}{c}{\bf{WS}} \\
\hline
\bf{IS$_{1-\alpha}$} &0.1 &0.3 &0.5 &0.7 &0.9 
\\ \hline
LDM &1.774 &2.147 &2.616 &3.298 &4.830 \\ 
CQLDM &1.770 &2.138 &2.592 &3.229 &4.507 \\ \hline
\bf{IW$_{1-\alpha}$} &0.1 &0.3 &0.5 &0.7 &0.9 
\\ \hline
LDM &0.190 &0.583 &1.022 &1.575 &2.506 \\ 
CQLDM &0.209 &0.643 &1.132 &1.762 &2.899 \\ \hline
\end{tabular}
\end{center}
\end{table}
\begin{table}[t!]
\caption{Quantile Scores for $\gamma \in \Gamma$ ($\times$10$^{-2}$)}
\label{Tab_pinball}
\begin{center}
\small
\begin{tabular}{llllll}
\hline
\multicolumn{6}{c}{\bf{2mT}} \\
\hline
\bf{QS$_{\gamma}$} &0.05 &0.15 &0.25 &0.35 &0.45 
\\ \hline
LDM  &6.34 &12.96 &17.10 &19.61 &20.79\\ 
CQLDM  &5.93 &12.69 &16.94 &19.53 &20.74\\ \hline
\bf{QS$_{\gamma}$} &0.55 &0.65 &0.75 &0.85 &0.95 
\\ \hline
LDM  &20.73 &19.44 &16.83 &12.63 &6.05\\ 
CQLDM  &20.69 &19.37 &16.70  &12.39 &5.68\\ \hline
\hline
\multicolumn{6}{c}{\bf{WS}} \\
\hline
\bf{QS$_{\gamma}$} &0.05 &0.15 &0.25 &0.35 &0.45 
\\ \hline
LDM  &11.07 &23.59 &31.75 &36.97 &39.71 \\ 
CQLDM  &10.61 &23.34 &31.65 &36.93 &39.66\\ \hline
\bf{QS$_{\gamma}$} &0.55 &0.65 &0.75 &0.85 &0.95 
\\ \hline
LDM  &40.13  &38.18 &33.64 &25.88 &13.08\\ 
CQLDM  &40.01 &37.92 &33.15 &25.09 &11.92\\ \hline
\end{tabular}
\end{center}
\end{table}
The PICP results across the target coverage levels are shown in Figures \ref{PICP_2mT}-\ref{PICP_WS} for the 2mT and WS test sets, respectively. To better visualize deviations at lower $(1-\alpha)$ scales, plots of the percentage drifts from the nominal target, computed as \(\%[ \text{PICP}_{1-\alpha} - (1-\alpha)]/{(1-\alpha)}\), are also included.
\begin{figure*}[t!]
  \centering
  \includegraphics[width=0.98\textwidth]{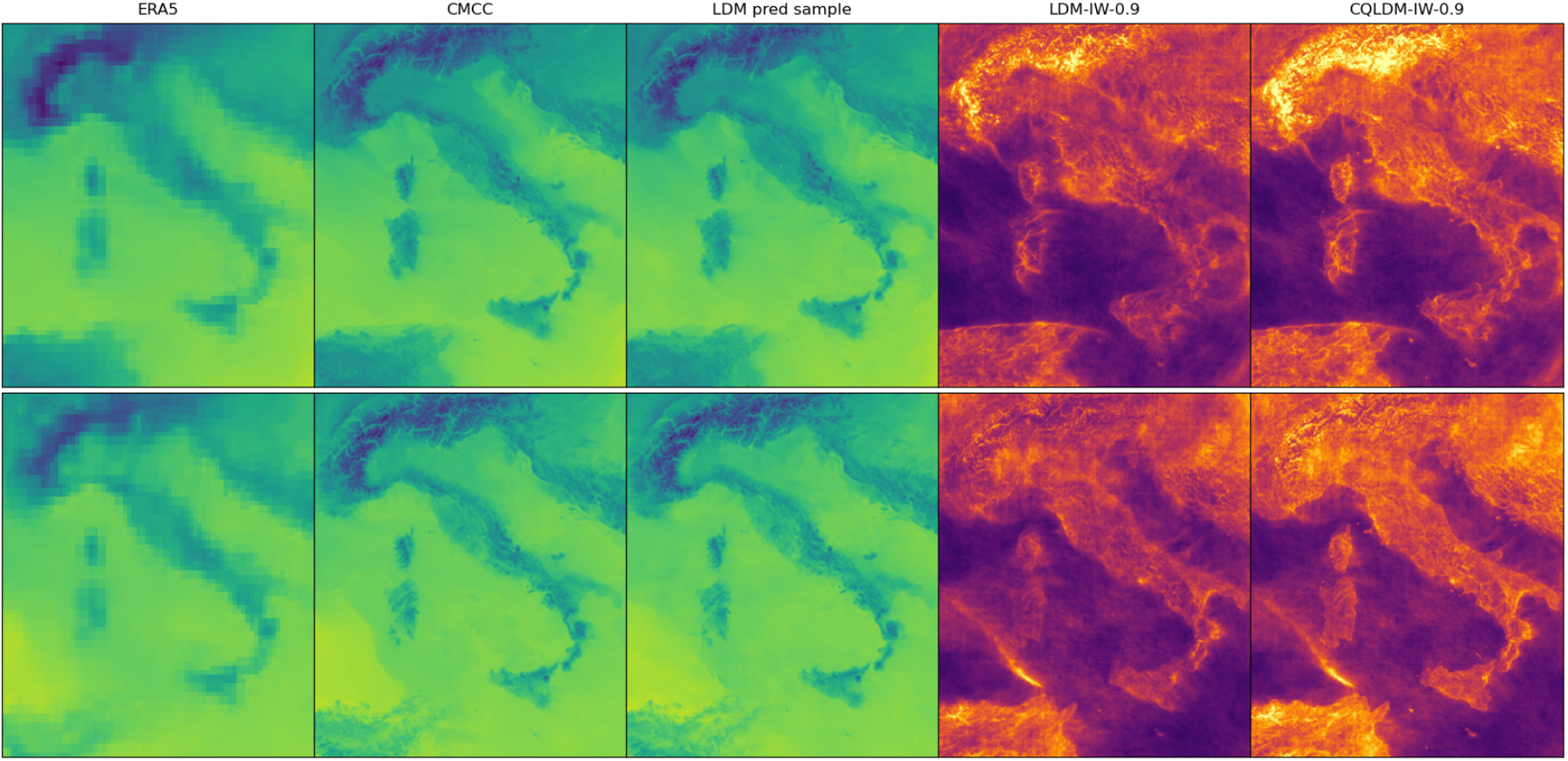}
  \caption{Samples of predictions from 2mT test set}
  \label{Fig-2mT_preds}
\end{figure*}
\begin{figure*}[t!]
  \centering
  \includegraphics[width=0.98\textwidth]{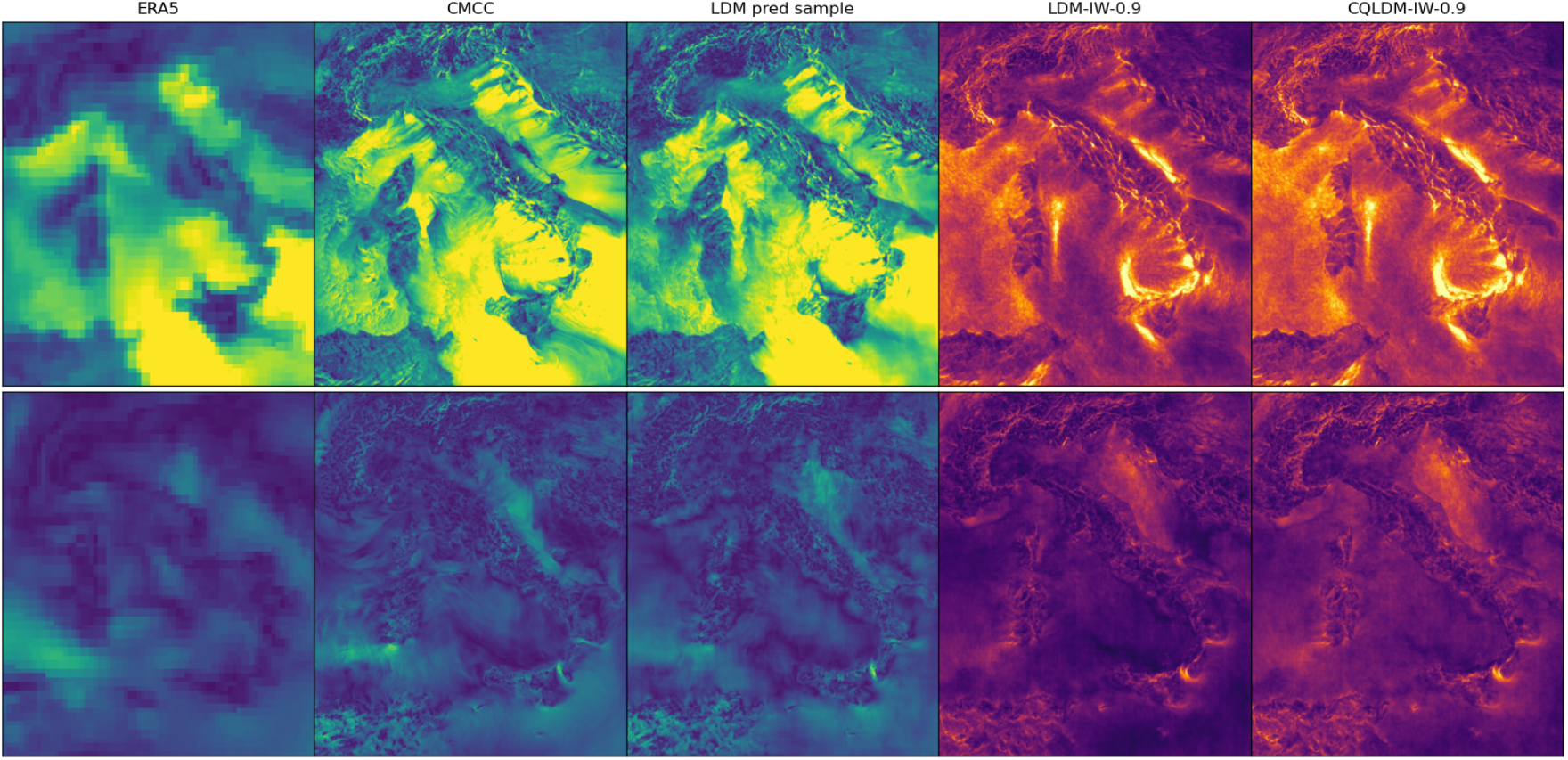}
  \caption{Samples of predictions from WS test set}
  \label{Fig-WS_preds}
\end{figure*}

The LDM baseline exhibits under-dispersed prediction intervals of comparable magnitude on both datasets, with varying degrees of miscoverage at the specified confidence thresholds. Such results confirm and complement the observations reported in previous works exploring residual corrective diffusion models for km-scale downscaling on different datasets (see \cite{Mardani2025}).
The grid-wise PICP analysis for 1-$\alpha$=0.9, shown in Figure \ref{fig:grid_picp_W}, reveals distinct spatial patterns among the variables. Notably, significant local 2mT reconstruction miscoverage appears over northern regions (e.g., the Alpine area) and along coastal boundaries. Conversely, WS shows more pronounced deviations in maritime regions beyond mountain ranges, as apparent, e.g., near Sicily, Sardinia, and in the northern Adriatic.

The introduction of the conformal inference framework (CQLDM) improves reliability across both datasets. Inspection of the grid-wise interval widths in Figure \ref{fig:grid_picp_W} demonstrates that local adaptations predominantly affect areas with poorer initial calibration, as expected. These corrections are quantitatively confirmed by increased average interval widths reported in Table \ref{Tab_IS_IW}. Similar behavior emerges in the sample-wise prediction analysis, as illustrated in Figures \ref{Fig-2mT_preds} and \ref{Fig-WS_preds}, which display prediction samples from the 2mT and WS test sets under various conditions.
Proper scoring rules further support the calibration findings. Stable results are observed for both the interval score (IS) and quantile score (QS) across confidence levels, with the CQLDM generally achieving lower values. The IS increases with 1-$\alpha$, reflecting wider intervals required to meet higher confidence targets. In contrast, the QS shows more elevated values near central quantiles, consistent with the percentage deviation of the PICP, possibly due to challenges in characterizing narrower intervals under volatile conditions.

Although the introduction of the conformal inference framework has substantially improved performance, there remains scope for further refinement. Conformal prediction fundamentally relies on the assumption of exchangeability among data points, which was implicitly accepted here through a straight temporal split of a reduced observation set. Further analysis of this assumption across multiple data splits exploited for conformity score computation relative to the test set is warranted. Such investigations may provide additional insights since violations of exchangeability - caused by factors like distributional shifts or temporal dependencies - can impact the principled guarantees and necessitate the use of additional techniques to address these challenges.
To balance computational costs associated with extensive diffusion sampling to derive conditional quantiles under limited GPU resources, this work employed a constrained calibration set size. Expanding the calibration base to better encompass test conditions would be a natural next step, followed by integration of methods to address potential distributional shifts. Furthermore, extending the conformal approach to multivariate settings may enable more precise local adjustments by leveraging spatial or temporal dependencies among related grid points. These improvements may be further complemented by incorporating temporal context within the overall network conditioning set.

\section{Conclusions and Next Developments}
This work targeted the challenging AI-enabled dynamical downscaling task, striving to match the capabilities of physics-based regional downscaling systems while enabling scalable sampling to support reliable uncertainty quantification. Focusing on state-of-the-art residual corrective diffusion models, we addressed the critical issue of overconfidence in the probabilistic reconstructions. To this end, we extended the downscaling framework by introducing a grid-wise conformalized quantile regression approach. Experiments conducted on the challenging task of downscaling ERA5 reanalyses to kilometer-scale reconstructions using the COSMO5.0 CLM9 model over Italy demonstrated that incorporating conformal inference yields grid-point-level uncertainty estimates with significantly improved reliability and stable probabilistic scores.

Several avenues for future research remain open. These include exploring multivariate conformal prediction methods and addressing potential violations of exchangeability in data splits. Further extensions could incorporate additional uncertainty quantification techniques to better characterize epistemic uncertainty in conditioned predictions, such as approximate Bayesian inference and deep ensembles. Applying the framework to neural network models that incorporate temporal context also represents a promising direction. A more thorough investigation of diffusion model hyperparameters and backbone architectural designs, particularly regarding their impact on sampling dispersion, is warranted. Additionally, we plan to experiment further dynamical downscaling applications.

\bibliography{mybibfile_v01}

\end{document}